\documentclass[10pt,twocolumn,letterpaper]{article}

\usepackage{cvpr}
\usepackage{times}
\usepackage{epsfig}
\usepackage{graphicx}
\usepackage{amsmath}
\usepackage{amssymb}
\usepackage{epstopdf}
\usepackage{multirow}
\usepackage{algorithm}
\usepackage{algpseudocode}
\usepackage{array}
\usepackage{graphicx}
\usepackage{float}
\usepackage{subcaption}
\usepackage{bbm} 
\usepackage[font=footnotesize,margin=2pt,skip=2pt]{caption}

\usepackage[pagebackref=true,breaklinks=true,letterpaper=true,colorlinks,bookmarks=false]{hyperref}
\usepackage[british,UKenglish,USenglish,english,american]{babel}

\cvprfinalcopy 

\def\cvprPaperID{2969} 

\ifcvprfinal\fi
\begin{document}

\title{Bidirectional Learning for Domain Adaptation of Semantic Segmentation}

\author{Yunsheng Li \thanks{This work was done when Yunsheng Li is an intern at Microsoft Cloud $\&$ AI}\\
UC San Diego\\
{\tt\small yul554@eng.ucsd.edu}
\and
Lu Yuan\\
Microsoft\\
{\tt\small luyuan@microsoft.com}
\and
Nuno Vasconcelos\\
UC San Diego\\
{\tt\small nvasconcelos@ucsd.edu}
}


\maketitle
\thispagestyle{empty}

\begin{abstract}
  Domain adaptation for semantic image segmentation is very necessary since manually labeling large datasets with pixel-level labels is expensive and time consuming. Existing domain adaptation techniques either work on limited datasets, or yield not so good performance compared with supervised learning. In this paper, we propose a novel bidirectional learning framework for domain adaptation of segmentation. Using the bidirectional learning, the image translation model and the segmentation adaptation model can be learned alternatively and promote to each other. Furthermore, we propose a self-supervised learning algorithm to learn a better segmentation adaptation model and in return improve the image translation model. Experiments show that our method is superior to the state-of-the-art methods in domain adaptation of segmentation with a big margin. The source code is available at \href{https://github.com/liyunsheng13/BDL}{https://github.com/liyunsheng13/BDL}.
\end{abstract}

\section{Introduction}

Recent progress on image semantic segmentation~\cite{long2015fully} has been driven by deep neural networks trained on large datasets. Unfortunately, collecting and manually annotating large datasets with dense pixel-level labels has been extremely costly due to large amount of human effort is required. Recent advances in computer graphics make it possible to train CNNs on photo-realistic synthetic images with computer-generated annotations~\cite{richter2016playing,ros2016synthia}. Despite this, the domain mismatch between the real images (\emph{target}) and the synthetic data (\emph{source}) cripples the models' performance.
Domain adaptation addresses this domain shift problem. Specifically, we focus on the hard case of the problem where no labels from the target domain are available. This class of techniques is commonly referred to as Unsupervised Domain Adaptation.

Traditional methods for domain adaptation involve minimizing some measure of distance between the source and the target distributions. Two commonly used measures are the first and second order moment~\cite{carlucci2017autodial}, and learning the distance metrics using Adversarial approaches~\cite{tzeng2017adversarial,vu2017domain}. Both approaches have had good success in the classification problems (\eg, MNIST~\cite{lecun1998gradient}, USPS~\cite{friedman2001elements} and SVHN~\cite{netzer2011reading}); however, as pointed out in~\cite{zhang2017curriculum}, their performance is quite limited on the semantic segmentation problem.

Recently, domain adaptation for semantic segmentation has made good progress by separating it into two sequential steps. It firstly translates images from the source domain to the target domain with an image-to-image translation model (\eg, CycleGAN~\cite{zhu2017unpaired}) and then add a discriminator on top of the features of the segmentation model to further decrease the domain gap~\cite{hoffman2017cycada,wu2018dcan}. When the domain gap is reduced by the former step, the latter one is easy to learn and can further decrease the domain shift. Unfortunately, the segmentation model very relies on the quality of image-to-image translation. Once the image-to-image translation fails, nothing can be done to make it up in the following stages.

In this paper, we propose a new \emph{bidirectional learning} framework for domain adaptation of image semantic segmentation. The system involves two separated modules: image-to-image translation model and segmentation adaptation model similar to~\cite{hoffman2017cycada,wu2018dcan}, but the learning process involves two directions (\ie, ``translation-to-segmentation" and ``segmentation-to-translation"). The whole system forms a closed-loop learning. Both models will be motivated to promote each other alternatively, causing the domain gap to be gradually reduced. Thus, how to allow one of both modules providing positive feedbacks to the other is the key to success.

On the forward direction (\ie, ``translation-to-segmentation", similar to~\cite{hoffman2017cycada,wu2018dcan}), we propose a \emph{self-supervised learning} (SSL) approach in training our segmentation adaptation model. Different from segmentation models trained on real data, the segmentation adaptation model is trained on both synthetic and real datasets, but the real data has no annotations. At every time, we may regard the predicted labels for real data with high confidence as the approximation to the ground truth labels, and then use them only to update the segmentation adaptation model while excluding predicted labels with low confidence. This process is referred as \emph{self-supervised learning}, which aligns two domains better than one-trial learning that is widely used in existing approaches. Furthermore, better segmentation adaptation model would contribute to better translation model through our backward direction learning.

On the backward direction (\ie, ``segmentation-to-translation"), our translation model would be iteratively improved by the segmentation adaptation model, which is different from~\cite{hoffman2017cycada,wu2018dcan} where the image-to-image translation is not updated once the model is trained. For the purpose, we propose a new \emph{perceptual loss}, which forces the semantic consistency between every image pixel and its translated version, to build the bridge between translation model and segmentation adaptation model. With the constraint in the translation model, the gap in visual appearance (\eg, lighting, object textures), between the translated images and real datasets (\emph{target}) can be further decreased. Thus, the segmentation model can be further improved through our forward direction learning.

From the above two directions, both the translation model and the segmentation adaptation model complement each other, which helps achieve state-of-the-art performance in adapting large-scale rendered image dataset SYNTHIA~\cite{ros2016synthia}/GTA5~\cite{richter2016playing}, to real image dataset, Cityscapes~\cite{cordts2016cityscapes}, and outperform other methods by a large margin. Moreover, the proposed method is general to different kinds of backbone networks.

In summary, our key contributions are: \vspace{-0.3em}
\begin{enumerate}
  \item We present a \emph{bidirectional learning} system for semantic segmentation, which is a closed loop to learn the segmentation adaptation model and the image translation model alternatively.\vspace{-0.3em}
  \item We propose a \emph{self-supervised learning} algorithm for the segmentation adaptation model, which incrementally align the source domain and the target domain at the feature level, based on the translated results.\vspace{-0.3em}
  \item We introduce a new \emph{perceptual loss} to the image-to-image translation, which supervises the translation by the updated segmentation adaptation model. 
\end{enumerate}

\section{Related Work}

\paragraph{Domain Adaptation.} When transferring knowledge from virtual images to real photos, it is often the case that there exists some discrepancy from the training to the test stage. Domain adaptation aims to rectify this mismatch and tune the models toward better generalization at testing~\cite{patel2015visual}. The existing work on domain adaptation has mainly  focused on image classification~\cite{saenko2010adapting}. A lot of work aims to learn domain-invariant representations through minimizing the domain distribution discrepancy. Maximum Mean Discrepancy (MMD) loss~\cite{geng2011daml}, computing the mean of representations, is a common distance metric between two domains. As the extension to MMD, some statistics of feature distributions such as mean and covariance~\cite{carlucci2017autodial,mancini2018boosting} are used to match two different domains. Unfortunately, when the distribution is not Gaussian, solely matching mean and covariance is not enough to align the two different domains well.

Adversarial learning~\cite{goodfellow2014generative} recently becomes popular, and another kind of domain adaptation methods. It reduces the domain shift by forcing the features from different domains to fool the discriminator. \cite{tzeng2017adversarial} would be the pioneer work, which introduces an adversarial loss on top of the high-level features of the two domains with the classification loss for the source dataset and achieves a better performance than the statistical matching methods. Expect for adversarial loss, some work proposed some extra loss functions to further decrease the domain shift, such as reweighted function for each class~\cite{chen2018re}, and disentangled representations for separated matching~\cite{vu2017domain}. All of these methods work on simple and small classification datasets (\eg, MNIST~\cite{lecun1998gradient} and SVHN~\cite{netzer2011reading}), and may have quite limited performance in more challenging tasks, like segmentation. \vspace{-0.7em}

\paragraph{Domain Adaptation for Semantic Segmentation.} Recently, more domain adaptation techniques are proposed for semantic segmentation models, since an enormous amount of labor-intensive work is required to annotate so many images that are needed to train high-quality segmentation networks. A possible solution to alleviate the human efforts is to train networks on virtual data which is labeled automatically. For example, GTA5~\cite{richter2016playing} and SYHTHIA~\cite{ros2016synthia} are two popular synthetic datasets of city streets with overlapped categories, similar views to the real datasets (\eg, CITYSCAPE~\cite{cordts2016cityscapes}, CamVid~\cite{BrostowSFC:ECCV08}). Domain adaptation can be used to align the synthetic and the real datasets.

The first work to introduce domain adaptation for semantic segmentation is~\cite{hoffman2016fcns}, which does the global and local alignments between two domains in the feature level. Curriculum domain adaptation~\cite{zhang2017curriculum} estimates the global distribution and the labels for the superpixel, and then learns a segmentation model for the finer pixel. In ~\cite{tsai2018learning}, multiple discriminators are used for different level features to reduce domain discrepancy. In~\cite{saleh2018effective}, foreground and background classes are separately treated for decreasing the domain shift respectively. All these methods target to directly align features between two domains. Unfortunately, the visual (\eg, appearance, scale, etc.) domain gap between synthetic and real data usually makes it difficult for the network to learn transferable knowledge.

Motivated by the recent progress of unpaired image-to-image translation work (\eg, CycleGAN~\cite{zhu2017unpaired}, UNIT~\cite{liu2017unsupervised}, MUNIT~\cite{huang2018multimodal}), the mapping from virtual to realistic data is regarded as the image synthesis problem. It can help reduce the domain discrepancy before training the segmentation models. Based on the translated results, Cycada~\cite{hoffman2017cycada} and DCAN~\cite{wu2018dcan} further align features between two domains in feature level. By separately reducing the domain shift in learning,  these approaches obtained the state-of-the-art performance. However, the performance is limited by the quality of image-to-image translation. Once it fails, nothing can be done in the following step. To address this problem, we introduce a bidirectional learning framework where both translation and segmentation adaption models can promote each other in a closed loop.

There are two most related work. In \cite{dundar2018domain}, the segmentation model is also used to improve the image translation, but not to adapt the source domain to the target domain since it is only trained on source data. \cite{zou2018unsupervised} also proposed a self-training method for training the segmentation model iteratively. However, the segmentation model is only trained on source data and uses none of image translation techniques.\vspace{-0.6em}


\paragraph{Bidirectional Learning.} The kind of techniques were first proposed to solve the neural machine translation problem, such as~\cite{he2016dual,niu2018bi}, which train a language translation model for both directions of a language pair. It improves the performance compared with the uni-direction learning and reduces the dependency on large amount of data. Bidirectional learning techniques were also extended to image generation problem~\cite{pontes2018bidirectional}, which trains a single network for both classification and image generation problem from both top-to-down and down-to-top directions. A more related work~\cite{russo2017source} proposed bidirectional image translation (\ie, source-to-target, and target-to-source), then trained two classifiers on both domains respectively and finally fuses the classification results. By contrast, our bidirectional learning refers to translation boosting the performance of segmentation and vise verse. The proposed method is used to deal with the semantic segmentation task.

\section{Method}

Given the source dataset $\mathcal{S}$ with segmentation labels $Y_{\mathcal{S}}$ (\eg, synthetic data generated by computer graphics) and the target dataset $\mathcal{T}$ with no labels (\ie, real data), we want to train a network for semantic segmentation, which is finally tested on the target dataset $\mathcal{T}$. Our goal is to make its performance to be as close as possible to the model trained on $\mathcal{T}$ with ground truth labels $Y_{\mathcal{T}}$. The task is unsupervised domain adaptation for semantic segmentation. The task is not easy since the visual (\eg, lighting, scale, object textures, etc.) domain gap between $\mathcal{S}$ and $\mathcal{T}$ makes it difficult for the network to learn transferable knowledge at once.

To address this problem, the recent work~\cite{hoffman2017cycada} proposed two separated subnetworks. One is  image-to-image translation subnetwork $\bf{F}$ which learn to translate an image from  $\mathcal{S}$ to $\mathcal{T}$ in absence of paired examples. The another is segmentation adaptation subnetwork $\bf{M}$ that is trained on translated results $\bf{F}(\mathcal{S})$, which have the same labels $Y_{\mathcal{S}}$ to $\mathcal{S}$, and the target images $\mathcal{T}$ with no labels. Both subnetworks are learnt in a sequential way shown in Figure~\ref{Fig:feedback_f_m}(a). Such a two-stage solution has two advantages: 1) $\bf{F}$ helps decrease the visual domain gap; 2) when domain gap is reduced, $\bf{M}$ is easy to learn, causing better performance. However, the solution has some limitations. Once $\bf{F}$ is learnt, it is fixed. There is no feedback from $\bf{M}$ to boost the performance of $\bf{F}$. Besides, one-trial learning for $\bf{M}$ seems to just learn limited transferable knowledge.

In this section, we propose a new learning framework which can address the above two issues well. We inherit the way of separated subnetworks, but employ a \emph{bidirectional learning} instead (in Section~\ref{sec:iter_tr}), which uses a closed-loop to iteratively update both $\bf{F}$ and $\bf{M}$. Furthermore, we introduce a $\emph{self-supervised learning}$ to allow $\bf{M}$ being self-motivated in training (in Section~\ref{sec:SSL}). The network architecture and loss functions are presented in Section~\ref{sec:feat_align}.

\subsection{Bidirectional Learning}
\label{sec:iter_tr}

\begin{figure}[t] 
    \setlength{\belowcaptionskip}{-0.3cm}
    \centering
    \includegraphics[width=.98\linewidth]{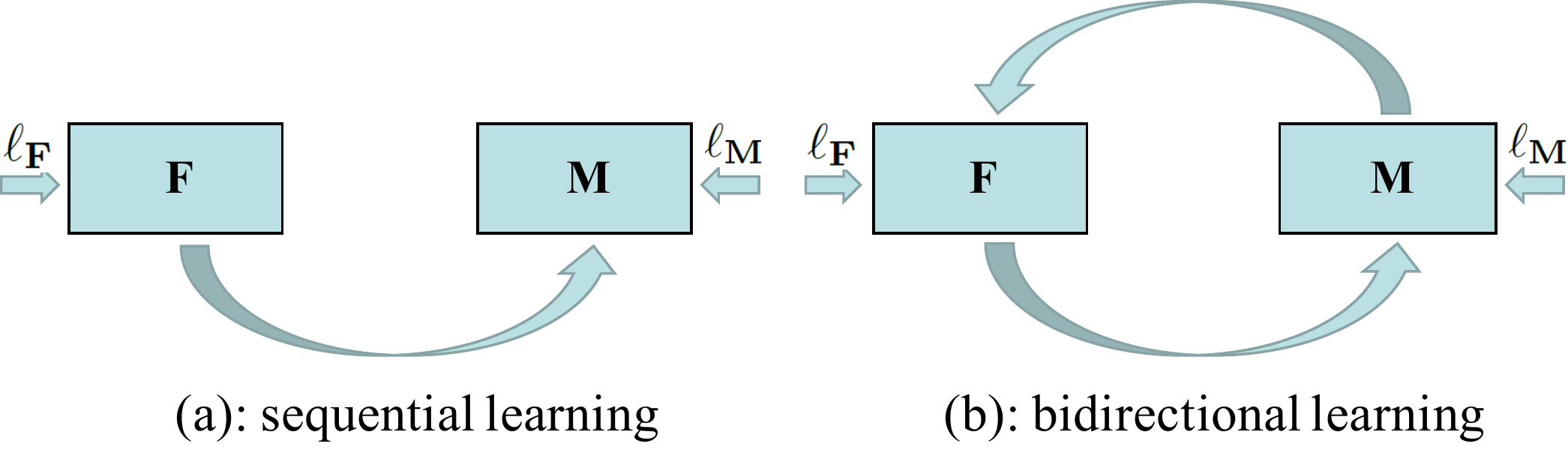} 
    \caption{Sequential Learning vs Bidirectional Learning}
    \label{Fig:feedback_f_m} 
    \vspace{-1em}
\end{figure}
Our learning consists of two directions shown in Figure~\ref{Fig:feedback_f_m}(b). 

The forward direction (\ie, ${\bf{F}}\rightarrow {\bf{M}}$) is similar to the behavior of previous sequential learning~\cite{hoffman2017cycada}. We first train the image-to-image translation model $\bf{F}$ using images from $\mathcal{T}$ and $\mathcal{S}$. Then, we get the translated results ${\mathcal{S}'}=\bf{F}(\mathcal{S})$. Note that $\bf{F}$ won't change the labels of ${\mathcal{S}'}$, which are the same to $Y_{\mathcal{S}}$ (labels of $\mathcal{S}$). Next, we train the segmentation adaptation model $\bf{M}$ using $\mathcal{S}'$ with $Y_{\mathcal{S}}$ and $\mathcal{T}$. The loss function to learn $\bf{M}$ can be defined as: 
\begin{equation}
\label{equ:seg_loss}
\ell_{\bf{M}} = \lambda_{adv}\ell_{adv}({\bf{M}}(\mathcal{S}'), {\bf{M}}(\mathcal{T})) + \ell_{seg}({\bf{M}}({\mathcal{S}'}), Y_{\mathcal{S}}),
\end{equation}
where $\ell_{adv}$ is adversarial loss that enforces the distance between the feature representations of $\mathcal{S}'$ and the feature representations of $\mathcal{T}$ (obtained after $\mathcal{S}'$, $\mathcal{T}$ are fed into $\bf{M}$) as small as possible. $\ell_{seg}$ measures the loss of semantic segmentation. Since only ${\mathcal{S}'}$ have the labels, we solely measure the accuracy for the translated source images $\mathcal{S}'$.

The backward direction (\ie, $\bf{M} \rightarrow \bf{F}$) is newly added. The motivation is to promote $\bf{F}$ using updated $\bf{M}$. In~\cite{vu2017domain, huang2018multimodal}, a perceptual loss, which measures the distance of features obtained from a pre-trained network on object recognition, is used in the image translation network to improve the quality of translated result. Here, we use $\bf{M}$ to compute features for measuring the perceptual loss. By adding the other two losses: GAN loss and image reconstruction loss, the loss function for learning $\bf{F}$ can be defined as:
\begin{equation}
\label{equ:trans_loss}
\begin{aligned}
\ell_{\bf{F}} &= \lambda_{GAN} [\ell_{GAN}(\mathcal{S}', \mathcal{T}) + \ell_{GAN}(\mathcal{S}, \mathcal{T}')] \\
&+ \lambda_{recon} [\ell_{recon}(\mathcal{S}, {{\bf{F}}^{-1}}(\mathcal{S}')) + \ell_{recon}(\mathcal{T}, {\bf{F}}(\mathcal{T}')]\\
&+ \ell_{per}({\bf{M}}(\mathcal{S}), {\bf{M}}(\mathcal{S}')) +
\ell_{per}({\bf{M}}(\mathcal{T}), {\bf{M}}(\mathcal{T}'),
\end{aligned}
\end{equation}
where three losses are computed symmetrically, \ie, $\mathcal{S} \rightarrow \mathcal{T}$ and $\mathcal{T} \rightarrow \mathcal{S}$, to ensure the image-to-image translation consistent. The GAN loss $\ell_{GAN}$ enforces two distributions between $\mathcal{S}'$ and $\mathcal{T}$ similar to each other. $\mathcal{T}'={\bf{F}}^{-1}(\mathcal{T})$, where ${\bf{F}}^{-1}$ is the reverse function of $\bf{F}$ that maps the image from $\mathcal{T}$ to $\mathcal{S}$. The loss $\ell_{recon}$ measures the reconstruction error when the image from $\mathcal{S}'$ is translated back to $\mathcal{S}$. $\ell_{per}$ is the perceptual loss that we propose to maintain the semantic consistency between $\mathcal{S}$ and $\mathcal{S}'$ or between $\mathcal{T}$ and $\mathcal{T}'$. That is, once we obtained an ideal segmentation adaptation model $\bf{M}$, whether $\mathcal{S}$ and $\mathcal{S}'$, or $\mathcal{T}$ and $\mathcal{T}'$ should have the same labels, even although there is the visual gap between $\mathcal{S}$ and $\mathcal{S}'$, or between $\mathcal{T}$ and $\mathcal{T}'$.

\subsection{Self-supervised Learning for Improving $\bf{M}$}
\label{sec:SSL}

In the forward direction (\ie, ${\bf{F}}\rightarrow {\bf{M}}$), if the label is available for both the source domain $\mathcal{S}$ and the target domain $\mathcal{T}$, the fully supervised segmentation loss $\ell_{seg}$ is always the best choice to reduce the domain discrepancy. But in our case, the label for the target dataset is missing. As we known, self-supervised learning (SSL) has been used in semi-supervised learning before, especially when the labels of dataset are insufficient or noisy. Here, we use SSL to help promote the segmentation adaptation model $\bf{M}$.

Based on the prediction probability of $\mathcal{T}$, we can obtain some pseudo labels $\widehat Y_{\mathcal{T}}$ with high confidence. Once we have the pseudo labels, the corresponding pixels can be aligned directly with $\mathcal{S}$ according to the segmentation loss. Thus, we modify the overall loss function used to learn $\bf{M}$ (in Equation~\ref{equ:seg_loss}) as:
\begin{equation}
\label{equ:trans_loss_D_c}
\begin{aligned}
\ell_{\bf{M}} &= \lambda_{adv}\ell_{adv}({\bf{M}}(\mathcal{S}'), {\bf{M}}(\mathcal{T})) \\ 
&+ \ell_{seg}({\bf{M}}({\mathcal{S}'}), Y_{\mathcal{S}}) + \ell_{seg}({\bf{M}}({\mathcal{T}_{ssl}}), \widehat Y_{\mathcal{T}}),
\end{aligned}
\end{equation}
where $\mathcal{T}_{ssl} \subset \mathcal{T}$ is a subset of the target dataset in which the pixels have the pseudo labels $\widehat Y_{\mathcal{T}}$. It can be empty at the beginning. When a better segmentation adaptation model $\bf{M}$ is achieved, we can use $\bf{M}$ to predict more high-confident labels for $\mathcal{T}$, causing the size of $\mathcal{T}_{ssl}$ to grow. The recent work \cite{zou2018unsupervised} also use SSL for segmentation adaptation. By contrast, SSL used in our work is combined with adversarial learning, which can work much better for the segmentation adaptation model.

\begin{figure}[t]
    \setlength{\belowcaptionskip}{-0.3cm}
    \centering
    \includegraphics[width=.98\linewidth]{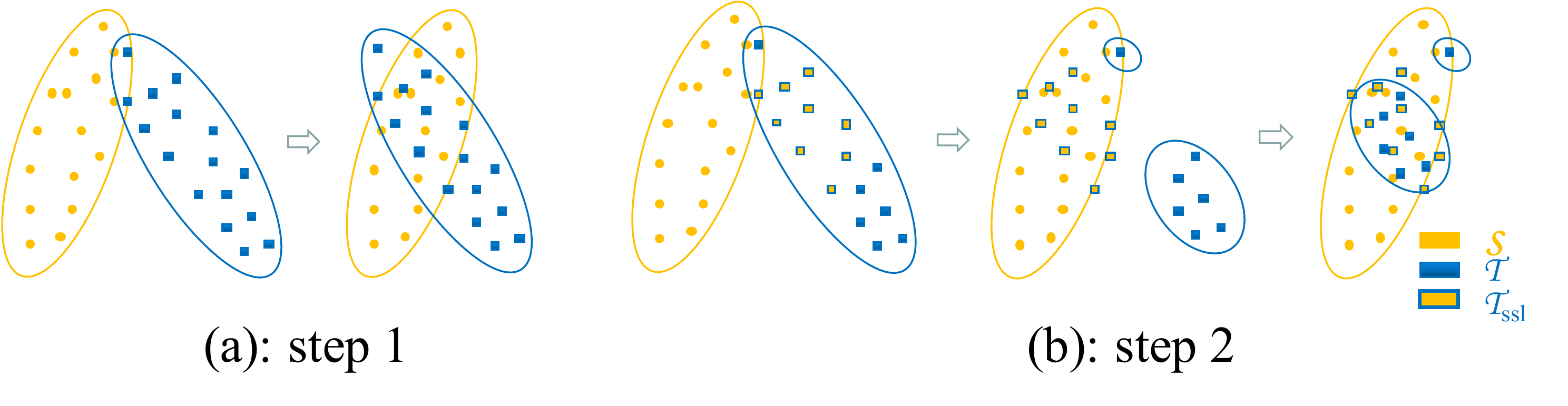}
    \caption{Self-supervised learning process}
    \label{Fig:ST_step_1_2}
\end{figure}

We use the illustration (shown in Figure \ref{Fig:ST_step_1_2}) to explain the principle of this process. When we learn the segmentation adaptation model for the first time, $\mathcal{T}_{ssl}$ is empty and the domain gap between $\mathcal{S}$ and $\mathcal{T}$ can be reduced with the loss shown in Equation \ref{equ:seg_loss}. This process is shown in Figure \ref{Fig:ST_step_1_2} (a). Then we pick up the points in the target domain $\mathcal{T}$  that have been well aligned with $\mathcal{S}$ to construct the subset $\mathcal{T}_{ssl}$. In the second step, we can easily shift $\mathcal{T}_{ssl}$ to $\mathcal{S}$ and keep them being aligned with the help of the segmentation loss provided by the pseudo labels. This process is shown in the middle of Figure \ref{Fig:ST_step_1_2} (b). Therefore, the amount of data in $\mathcal{T}$ that needs to be aligned with $\mathcal{S}$ is decreased. We can continue to shift the remaining data to $\mathcal{S}$ same as step 1, as shown the right side of Figure \ref{Fig:ST_step_1_2} (b). It worth noting that SSL helps adversarial learning process focus on the rest data that is not fully aligned at each step, since $\ell_{adv}$ can hardly change the data from $\mathcal{S}$ and $\mathcal{T}_{ssl}$ that has been aligned well.

\begin{algorithm}[t]
\caption{Training process of our network}\label{algr:train_proc}
\begin{algorithmic}
\Require{($\mathcal{S}$, $Y_\mathcal{S}$), ($\mathcal{T}$, $\mathcal{T}_{ssl}=\emptyset$), $\bf{M}^{(0)}$}
\Ensure{${\bf{M}}_N^{(K)}({{\bf{F}}^{(K)}})$}
\For{$k \gets 1$ to $K$ } (Bidirectional Learning)\\
    \hspace{\algorithmicindent}{train ${\bf{F}}^{(k)}$ with Equation \ref{equ:trans_loss} }\\
    \hspace{\algorithmicindent}{train ${\bf{M}}_0^{(k)}$ with Equation \ref{equ:seg_loss} }
    \For{$i \gets 1$ to $N$ } (SSL)\\
        \hspace{\algorithmicindent}\hspace{\algorithmicindent}{update $\mathcal{T}_{ssl}$ with ${\bf{M}}_{i-1}^{(k)}$ }\\
        \hspace{\algorithmicindent}\hspace{\algorithmicindent}{train ${\bf{M}}_i^{(k)}$ again with Equation \ref{equ:trans_loss_D_c} }
    \EndFor
\EndFor
\end{algorithmic}
\end{algorithm}

\begin{figure*}[t]
\setlength{\belowcaptionskip}{-0.5cm}
  \begin{minipage}{0.98\linewidth}
    \centering
    \includegraphics[width=.95\linewidth]{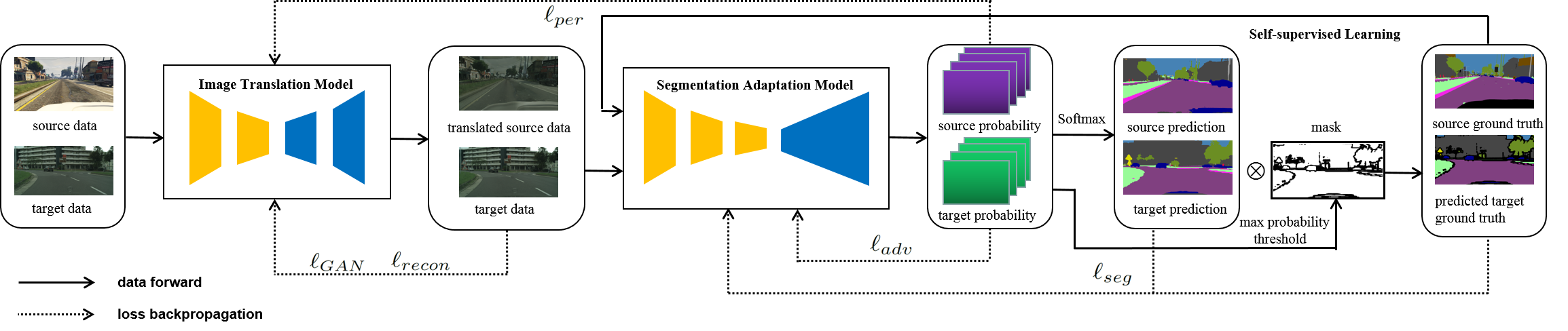}
    \caption{Network architecture and loss function}
    \label{Fig:network_arch}
  \end{minipage}
\end{figure*}
\begin{figure}[!htp]
\setlength{\belowcaptionskip}{-0.5cm}
  \begin{minipage}{0.98\linewidth}
    \centering
    \includegraphics[width=.95\linewidth]{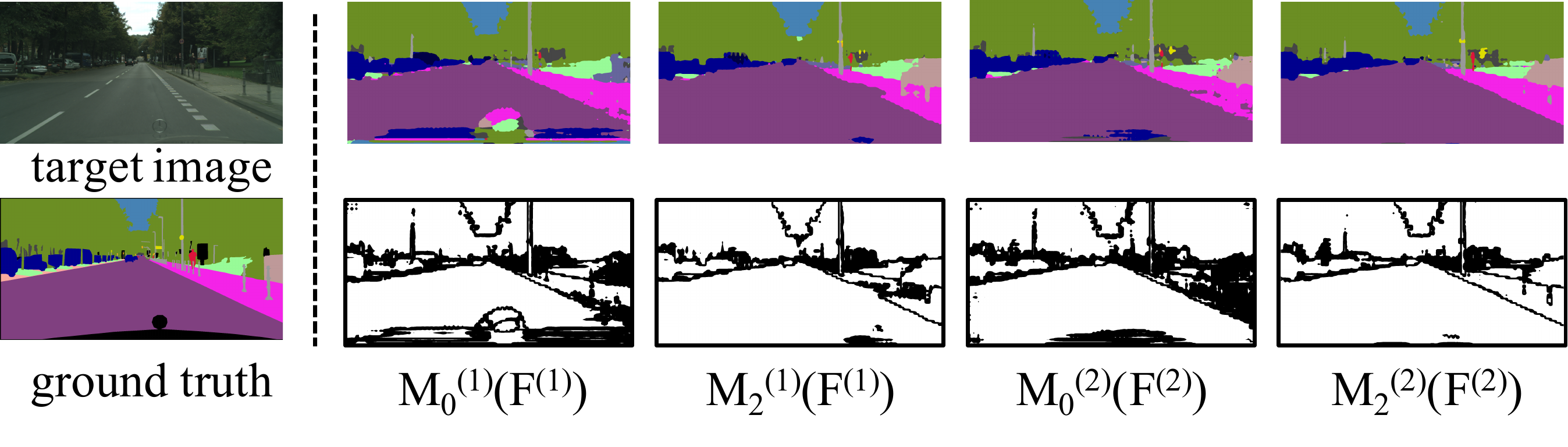}
    \caption{Segmentation result for each step in bidirectional learning}
    \label{Fig:seg_res}
  \end{minipage}
\end{figure}

\subsection{Network and Loss Function}
\label{sec:feat_align}
In this section, we introduce the network architecture (shown in Figure \ref{Fig:network_arch}), details of loss functions and the training process (shown in Algorithm \ref{algr:train_proc}).
The network is mainly composed with two components -- the image translation model and segmentation adaptation model. 

While the translation model is learned, the loss $\ell_{GAN}$ and loss $\ell_{recon}$ (shown in Figure \ref{Fig:network_arch} and Equation \ref{equ:trans_loss}) can be defined as:
$$
\ell_{GAN}(\mathcal{S}', \mathcal{T}) =  \mathbb{E}_{I_{\mathcal{T}}\sim \mathcal{T}}[D_{\bf{F}}(I_{\mathcal{T}})] + \mathbb{E}_{I_\mathcal{S}\sim \mathcal{S}}[1 - D_{\bf{F}}((I'_{\mathcal{S}}))],
$$ 
$$
\ell_{recon}(\mathcal{S}, {\bf{F}^{-1}}({\mathcal{S}'})) = \mathbb{E}_{I_\mathcal{S}\sim \mathcal{S}}[||{\bf{F}^{-1}}((I'_{\mathcal{S}}))-I_\mathcal{S}||_1], \quad\quad\;\;\;
$$
where $I_\mathcal{S}$ and $I_{\mathcal{T}}$ are the input images from source and target dataset. $I'_{\mathcal{S}}$ is the translated image given by $\bf{F}$. $D_{\bf{F}}$ is the discriminator added to reduce the difference between $I_{\mathcal{T}}$ and $I'_{\mathcal{S}}$. For the reconstruction loss, $L_1$ norm is used to keep the cycle consistency between $I_\mathcal{S}$ and ${\bf{F}}^{-1}(I'_{\mathcal{S}})$ when $\bf{F}^{-1}$ is the reverse function of $\bf{F}$. Here, we only show two losses for one direction, and $\ell_{GAN}(\mathcal{S}, \mathcal{T}'), \ell_{recon}(\mathcal{T}, {\bf{F}}({\mathcal{T}'}))$ can be defined similarly.

As shown in Figure \ref{Fig:network_arch}, the perceptual loss $\ell_{per}$ connects the translation model and segmentation adaptation model. When we learn the perceptual loss $\ell_{per}$ for the translation model, instead of only keeping the semantic consistency between $I_\mathcal{S}$ and its translated result $I'_\mathcal{S}$, we add another term weighted by $\lambda_{per\_{recon}}$, to keep the semantic consistency between $I_\mathcal{S}$ and its corresponding reconstruction ${\bf{F}}^{-1}(I'_\mathcal{S})$. With the new term, the translation model can be more stable especially for the reconstruction part. $\ell_{per}$ is defined as:
$$
    \begin{aligned}
        \ell_{per}({\bf{M}}(\mathcal{S}), {\bf{M}}(\mathcal{S}'))= \lambda_{per} \mathbb{E}_{I_\mathcal{S}\sim \mathcal{S}}||{\bf{M}}(I_\mathcal{S})-{\bf{M}}((I'_{\mathcal{S}}))||_1+ \quad\\
        \lambda_{per\_{recon}}\mathbb{E}_{I_\mathcal{S}\sim \mathcal{S}}[||{\bf{M}}({\bf{F}}^{-1}((I'_\mathcal{S})))-{\bf{M}}(I_\mathcal{S})||_1] \quad\quad
    \end{aligned}
    \label{equ:loss_perceptual}
$$
Due to the symmetry, $\ell_{per}({\bf{M}}(\mathcal{T}), {\bf{M}}(\mathcal{T}'))$ (shown in Equation \ref{equ:trans_loss}) can be defined in a similar way.

When the segmentation adaptation model is trained, it requires the adversarial learning with the loss $\ell_{adv}$ and the self-supervised learning with the loss $\ell_{seg}$ (shown in Equation \ref{equ:trans_loss_D_c}). For adversarial learning, we add a discriminator $D_{\bf{M}}$ to decrease the difference between the source and target probabilities shown in Figure \ref{Fig:network_arch}. $\ell_{adv}$ can be defined as:
$$
\begin{aligned}
\ell_{adv}({\bf{M}}(\mathcal{S}'), {\bf{M}}(\mathcal{T})) &=\mathbb{E}_{I_\mathcal{T}\sim \mathcal{T}}[D_{\bf{M}}({\bf{M}}(I_\mathcal{T}))] \quad\quad\quad\quad\quad\\
    &+\mathbb{E}_{I_\mathcal{S}\sim \mathcal{S}}[1 - D_{\bf{M}}({\bf{M}}(I'_\mathcal{S}))].
\end{aligned}
$$
The segmentation loss $\ell_{seg}$ uses the cross-entropy loss. For the source image $I_\mathcal{S}$, $\ell_{seg}$ can be defined as:
$$
\ell_{seg}({\bf{M}}({\mathcal{S}'}), Y_{\mathcal{S}})=-\frac{1}{HW}\sum_{H, W}\sum_{c=1}^{C}\mathbbm{1}_{[c=y_{\mathcal{S}}^{hw}]} \log P_{\mathcal{S}}^{hwc}, \quad\quad
$$
where $y_{\mathcal{S}}$ is the label map for $I_{\mathcal{S}}$, $C$ is the number of classes, $H$ and $W$ are the height and width of the output probability map. $P_{\mathcal{S}}$ is the source probability of the segmentation adaptation model which can be defined as $P_{\mathcal{S}}={\bf{M}}(I'_{\mathcal{S}})$. For the target image $I_{\mathcal{T}}$, we need to define how to choose the pseudo label map $\widehat y_{\mathcal{T}}$ for it. We choose to use a common method we call as "max probability threshold(MPT)" to filter the pixels with high prediction confidence in $I_{\mathcal{T}}$. Thus we can define $\widehat y_{\mathcal{T}}$ as $\widehat y_{\mathcal{T}}=\operatorname*{argmax}{\bf{M}}(I_{\mathcal{T}})$ and the mask map for $\widehat y_{\mathcal{T}}$ as $m_{\mathcal{T}}=\mathbbm{1}_{[\operatorname*{argmax}{\bf{M}}(I_{\mathcal{T}}) > \operatorname*{threshold}]}$. Thus the segmentation loss for $I_{\mathcal{T}}$ can be expressed as:
$$
    \ell_{seg}({\bf{M}}({\mathcal{T}_{ssl}}), \widehat Y_{\mathcal{T}})=-\frac{1}{HW}\sum_{H, W}m_{\mathcal{T}}^{hw}\sum_{c=1}^{C}\mathbbm{1}_{[c=y_{\mathcal{T}}^{hw}]} \log P_{\mathcal{T}}^{hwc},
$$
\noindent where $P_\mathcal{T}$ is the target output of $\bf{M}$.

We present the training processing in Algorithm \ref{algr:train_proc}. The training process consists of two loops. The outer loop is mainly to learn the translation model and the segmentation adaptation model through the forward direction and the backward direction. The inner loop is mainly used to implement the SSL process. In the following section, we will introduce how to choose the number of iteration for learning $\bf{F}$, $\bf{M}$, and how to estimate the MPT for SSL.

\section{Discussion}

To know the effectiveness of bidirectional learning and self-supervised learning for improving $\bf{M}$, we conduct some ablation studies. We use GTA5 \cite{richter2016playing} as the source dataset and Cityscapes \cite{cordts2016cityscapes} as the target dataset. The translation model is CycleGAN \cite{zhu2017unpaired} and the segmentation adaptation model is DeepLab V2 \cite{chen2018deeplab} with the backbone ResNet101~\cite{he2016deep}. All the following experiments use the same model, unless it is specified.

Here, we first provide the description of notations used in the following ablation study and tables. ${\bf{M}}^{(0)}$ is the initial model to start the bidirectional learning and is trained only with source data. ${\bf{M}}^{(1)}$ is trained with source and target data with adversarial learning. For ${\bf{M}}^{(0)}({\bf{F}}^{(1)})$, a translation model ${\bf{F}}^{(1)}$ is used to translate the source data and then a segmentation model ${\bf{M}}^{(0)}$ is learned based on the translated source data. ${\bf{M}}_{i}^{(k)}({\bf{F}}^{(k)})$ for $k=1,2$ and $i=0,1,2$ refers to the model of $k$-th iteration for the outer loop and $i$-th iteration for the inner loop in Algorithm \ref{algr:train_proc}.

\begin{table}[!hbt]
	\caption{Performance of bidirectinal learning}
	\centering
	\scriptsize
	\begin{tabular}{cc}
		\hline
		\multicolumn{2}{ c }{{ GTA5 $\rightarrow$ Cityscapes}} \\
		\hline
		model & mIoU\\
		\hline
		${\bf{M}}^{(0)}$ & 33.6\\
		\hline
		${\bf{M}}^{(1)}$ & 40.9\\
		\hline
		${\bf{M}}^{(0)}({\bf{F}}^{(1)})$ & 41.1\\
		\hline
		${\bf{M}}_0^{(1)}({\bf{F}}^{(1)})$ & 42.7\\
		\hline
		${\bf{M}}_0^{(2)}({\bf{F}}^{(2)})$ & 43.3\\
		\hline
	\end{tabular}
	\label{tab:perf_step_no_ST}
	\vspace{-1em}
\end{table}

\subsection{Bidirectional Learning without SSL}
\label{sec:role_iter_training}
We show the results obtained by the model trained in a bidirectional learning system without SSL. In Table \ref{tab:perf_step_no_ST}, ${\bf{M}}^{(0)}$ is our baseline model that gives the lowerbound for mIoU. We find a similar performance between the model ${\bf{M}}^{(1)}$ and ${\bf{M}}^{(0)}{(\bf{F}}^{(1)})$ both of which achieve more than $7\%$ improvement compared to ${\bf{M}}^{(0)}$ and about $1.6\%$ further improvement is given by ${\bf{M}}^{(1)}({\bf{F}}^{(1)})$. It means segmentation adaptation model and the translation model can work independently and when combined together which is basically one iteration of the bidirectional learning they can be complementary to each other. We further show that through continue training the bidirectional learning system, in which case ${\bf{M}}^{(1)}({\bf{F}}^{(1)})$ is used to replace ${\bf{M}}^{(0)}$ for the backward direction, a better performance can be given by the new model ${\bf{M}}_0^{(2)}({\bf{F}}^{(2)})$.

\begin{table*}[!hbt]
\scriptsize
\caption{\footnotesize{Performance of bidirectional learning with self-supervised learning}}
\centering
\setlength{\tabcolsep}{4pt}
\begin{tabular}{cccccccccccccccccccccc}
\hline
\multicolumn{22}{ c }{{\scriptsize GTA5 $\rightarrow$ Cityscapes}} \\
\hline
 & & \rotatebox{90}{road}  & \rotatebox{90}{sidewalk} &\rotatebox{90}{building} & \rotatebox{90}{wall} & \rotatebox{90}{fence} & \rotatebox{90}{pole} & \rotatebox{90}{t-light} & \rotatebox{90}{t-sign} & \rotatebox{90}{vegetation} & \rotatebox{90}{terrain} & \rotatebox{90}{sky} & \rotatebox{90}{person} & \rotatebox{90}{rider} & \rotatebox{90}{car} & \rotatebox{90}{truck} & \rotatebox{90}{bus} & \rotatebox{90}{train} & \rotatebox{90}{motorbike} & \rotatebox{90}{bicycle} & mIoU\\
\hline
  & ${\bf{M}}^{(0)}$ & 69.0	& 12.7 & 69.5 & 9.9 & 19.5 & 22.8 & 31.7 & 15.3 & 73.9 & 11.3 & 67.2 & 54.7 & 23.9 & 53.4 & 29.7 & 4.6 & 11.6 & 26.1 & 32.5 & 33.6 \\
\hline
\multirow{3}{*}{$k=1$}

 & ${\bf{M}}_0^{(1)}({\bf{F}}^{(1)})$ & 89.1 & 42.0 & 82.0 & 24.3 & 15.1 & 27.4 & 35.7 & 24.6 & 81.1 & 32.4 & 78.0 & 57.6 & 28.7 & 76.0 & 26.5 & 36.0 & 4.0 & 25.7 & 24.9 & 42.7 \\
\cline{2-22}
 & ${\bf{M}}_1^{(1)}({\bf{F}}^{(1)})$ & 91.2 & 47.8 & 84.0 & 34.8 & 28.9 & 31.7 & 37.7 & 36.0 & 84.0 & 40.4 & 76.6 & 57.9 & 25.3 & 80.4 & 31.2 & 41.7 & 2.8 & 27.2 & 32.4 & 46.8 \\
\cline{2-22}
 & ${\bf{M}}_2^{(1)}({\bf{F}}^{(1)})$ & 91.4 & 47.9 & 84.2 & 32.4 & 26.0 & 31.8 & 37.3 & 33.0 & 83.3 & 39.2 & 79.2 & 57.7 & 25.6 & 81.3 & 36.3 & 39.7 & 2.6 & 31.3 & 33.5 & 47.2 \\
\hline
\multirow{3}{*}{$k=2$}
  &${\bf{M}}_0^{(2)}({\bf{F}}^{(2)})$ & 88.2 & 41.3 & 83.2 & 28.8 & 21.9 & 31.7 & 35.2 & 28.2 & 83.0 & 26.2 & 83.2 & 57.6 & 27.0 & 77.1 & 27.5 & 34.6 & 2.5 & 28.3 & 36.1 & 44.3 \\
\cline{2-22}
  &${\bf{M}}_1^{(2)}({\bf{F}}^{(2)})$ & 91.2 & 46.1 & 83.9 & 31.6 & 20.6 & 29.9 & 36.4 & 31.9 & 85.0 & 39.7 & 84.7 & 57.5 & 29.6 & 83.1 & 38.8 & 46.9 & 2.5 & 27.5 & 38.2 & 47.6 \\
\cline{2-22}
  &${\bf{M}}_2^{(2)}({\bf{F}}^{(2)})$ & 91.0 & 44.7 & 84.2 & 34.6 & 27.6 & 30.2 & 36.0 & 36.0 & 85.0 & 43.6 & 83.0 & 58.6 & 31.6 & 83.3 & 35.3 & 49.7 & 3.3 & 28.8 & 35.6 & 48.5 \\
\hline
\end{tabular}
\label{tab:perf_step}
\vspace{-1em}
\end{table*}


\subsection{Bidirectional Learning with SSL}
\label{sec:role_of_sst}
In this section, we show how the SSL can further improve the ability of segmentation adaption model and in return influence the bidirectional learning process. In Table \ref{tab:perf_step}, we show results given by two iterations($k=1,2$) based on Algorithm \ref{algr:train_proc}. In Figure \ref{Fig:seg_res}, we show the segmentation results and the corresponding mask map given by the max probability threshold (MPT) which is $0.9$. In Figure \ref{Fig:seg_res}, the white pixels are the ones with prediction confidence higher than MPT and the black pixels are the low confident pixels. 

While $k=1$, when model ${\bf{M}}_0^{(1)}({\bf{F}}^{(1)})$ is updated to ${\bf{M}}_2^{(1)}({\bf{F}}^{(1)})$ with SSL, the mIoU can be improved by $4.5\%$.
We can find for each category when the IoU is below $50$, a big improvement can be got from ${\bf{M}}_0^{(1)}({\bf{F}}^{(1)})$ to ${\bf{M}}_2^{(1)}({\bf{F}}^{(1)})$. It can prove our previous analysis in section \ref{sec:SSL} that with SSL the well aligned data from source and target domain can be kept and the rest data can be further aligned through the adversarial learning process.


While $k=2$, we first replace ${\bf{M}}^{(0)}$ with ${\bf{M}}_2^{(1)}({\bf{F}}^{(1)})$ to start the backward direction. Without SSL the mIoU is $44.3$ which is a larger improvement compared to the results shown in Table \ref{tab:perf_step_no_ST}. It can further prove our discussion in section \ref{sec:role_iter_training} about the importance role played by the segmentation adaptation model in the backward direction. Furthermore, we can find from Table \ref{tab:perf_step}, although in the beginning of the second iteration the mIoU drops from $47.2$ to $44.3$, while SSL is induced, the mIoU can be promoted to $48.5$ which outperforms the results in the first iteration. From the segmentation results shown in Figure \ref{Fig:seg_res}, our findings can be further confirmed and the most important thing is as we improve the segmentation performance, the segmentation adaptation model can give more confident prediction which can be observed by the increasing white area in the mask map. It gives us the motivation to use the mask map to choose the threshold and number of iterations for the SSL process in Algorithm \ref{algr:train_proc}.

\begin{figure}[t]
    \begin{minipage}{0.48\linewidth}
            \scriptsize
            \centering
            \captionof{table}{Influence of threshold}
            \setlength{\tabcolsep}{4pt}
                \begin{tabular}{ccc}
                \hline
                \multicolumn{3}{ c }{{ GTA5 $\rightarrow$ Cityscapes}} \\
                \hline
                model & threshold   & mIoU\\
                \hline
                 ${\bf{M}}_1^{(1)}({\bf{F}}^{(1)})$ & $0.95$ & 45.7 \\
                \hline
                 ${\bf{M}}_1^{(1)}({\bf{F}}^{(1)})$ & $0.9$ & 46.8 \\
                \hline
                 ${\bf{M}}_1^{(1)}({\bf{F}}^{(1)})$ & $0.8$ & 46.4
                 \\
                \hline
                 ${\bf{M}}_1^{(1)}({\bf{F}}^{(1)})$ & $0.7$ & 45.9 \\
                \hline
                 ${\bf{M}}_1^{(1)}({\bf{F}}^{(1)})$ & $-$ & 44.9 \\
                \hline
                \end{tabular}
            \label{tab:thre_influ}
    \end{minipage}
    \begin{minipage}{0.48\linewidth}
        \scriptsize
        \centering
        \captionof{table}{Influence of N}
        \setlength{\tabcolsep}{4pt}
        \begin{tabular}{ccc}
        \hline
        \multicolumn{3}{ c }{{ GTA5 $\rightarrow$ Cityscapes}} \\
        \hline
        model & pixel ratio   & mIoU\\
        \hline
         ${\bf{M}}_0^{(1)}$ & $66\%$ & 40.9 \\
        \hline
         ${\bf{M}}_0^{(1)}({\bf{F}}^{(1)})$ & $69\%$ & 42.7 \\
         \hline
         ${\bf{M}}_1^{(1)}({\bf{F}}^{(1)})$ & $79\%$ & 46.8 \\
        \hline
         ${\bf{M}}_2^{(1)}({\bf{F}}^{(1)})$ & $81\%$ & 47.2
         \\
        \hline
         ${\bf{M}}_3^{(1)}({\bf{F}}^{(1)})$ & $81\%$ & 47.1 \\
        \hline
        \end{tabular}
        \label{tab:N_influ}
    \end{minipage}
\end{figure}

\subsection{Hyper Parameter Learning}
\begin{figure}[t]
\setlength{\belowcaptionskip}{-0.47cm}
  \begin{minipage}{0.98\linewidth}
    \includegraphics[width=.49\linewidth, height=0.4\linewidth]{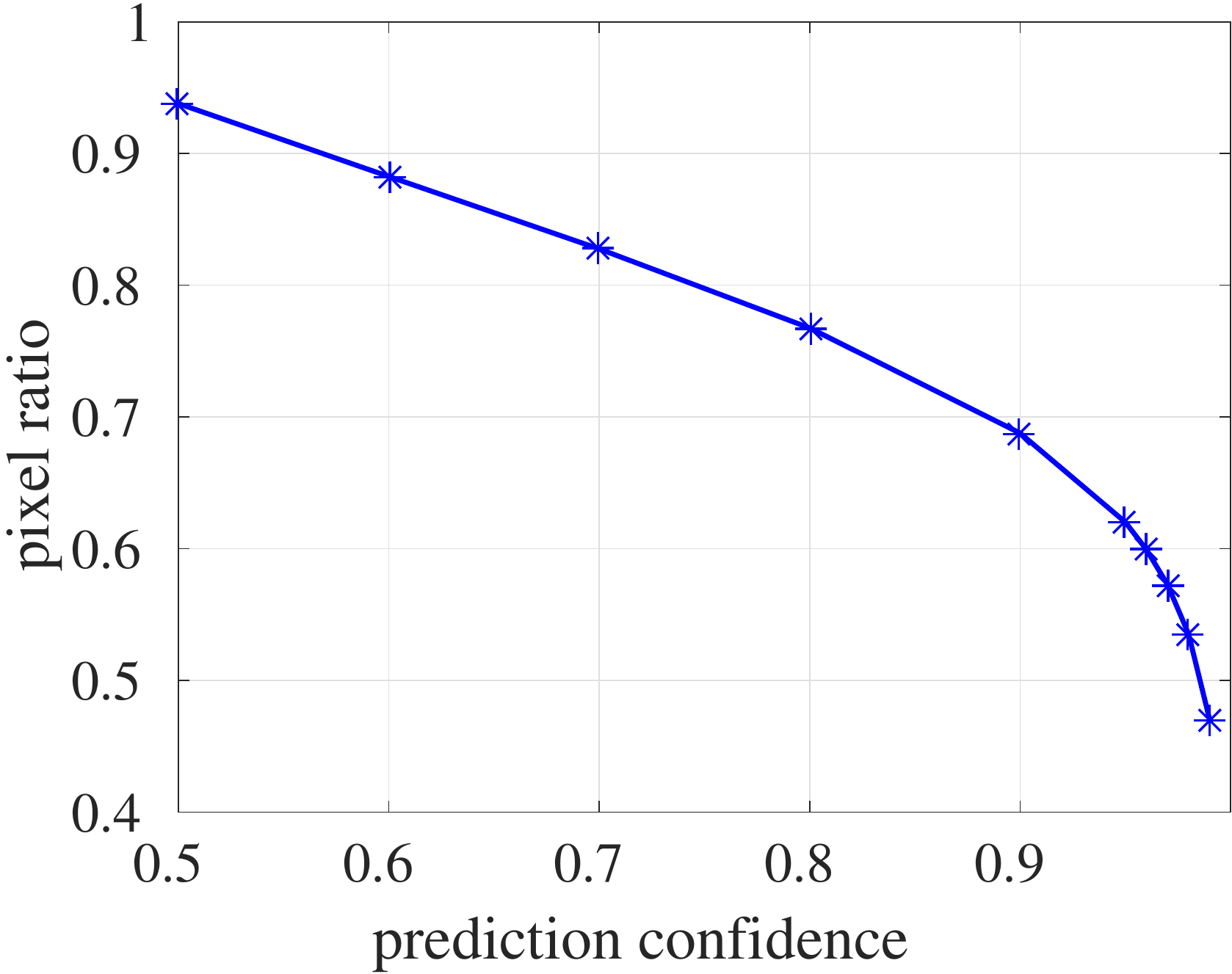} 
    \includegraphics[width=.49\linewidth]{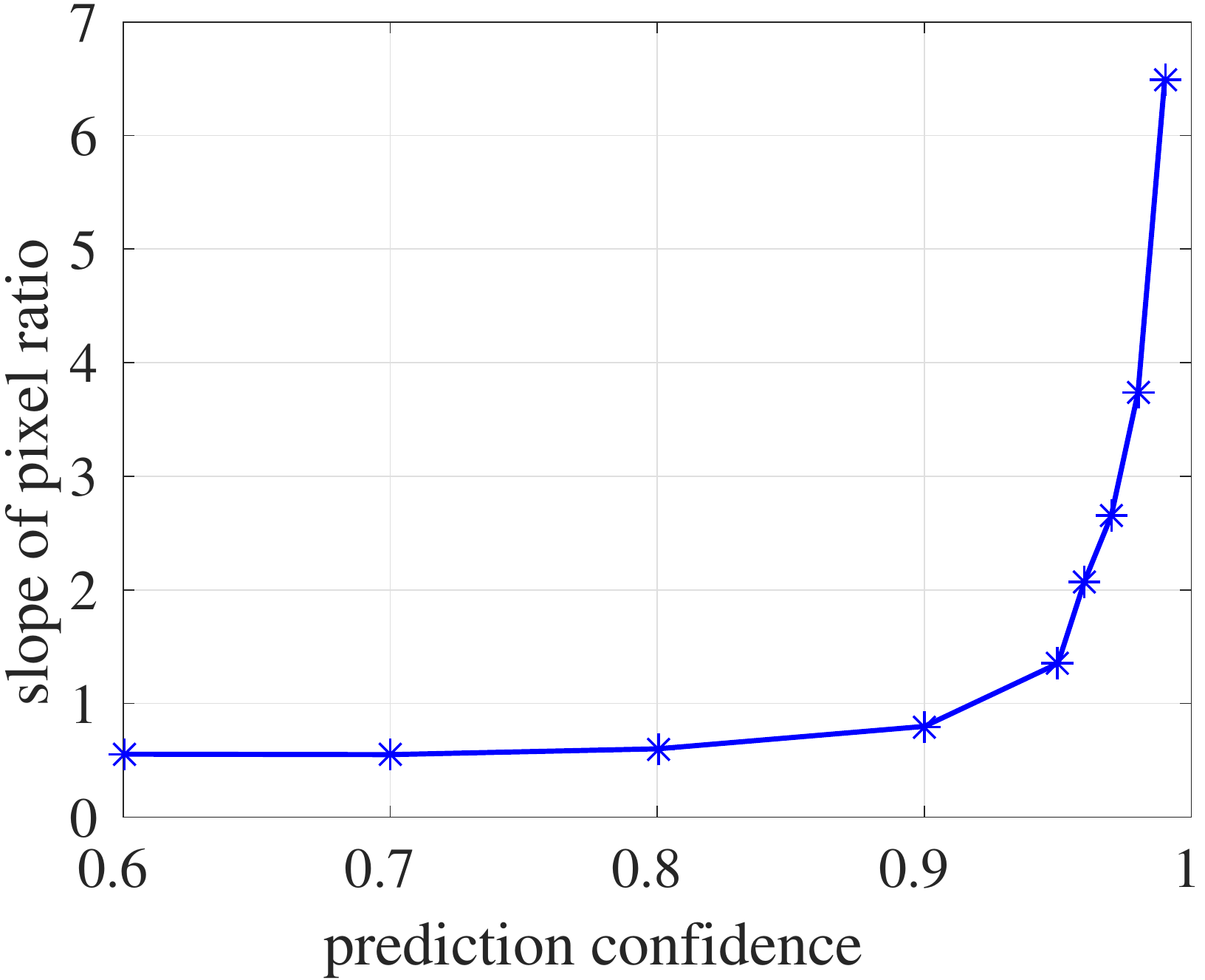} 
    \caption{Relationship between pixel ratio and the prediction confidence}
    \label{fig:ratio}
  \end{minipage} 
\end{figure}
We will describe how to choose the threshold to filter out data with high confidence and the iteration number $N$ in Algorithm \ref{algr:train_proc}. 

When we choose the threshold, we have to balance between two folds. On one hand, we desire the predicted labels with high confidence as many as possible (presented as white areas in Figure \ref{Fig:seg_res}). On the other hand, we want to avoid inducing too much noise caused by the incorrect prediction, namely, the threshold should be as high as possible. We present the relationship of the prediction confidence (maximum class probability of per pixel from $\bf{M}$) and the ratio between selected pixels and all pixels (\ie, percentage of all white areas shown in Figure \ref{Fig:seg_res}) on the left side of Figure \ref{fig:ratio}, then show the slope in the right side of Figure \ref{fig:ratio}. 
We can find when the prediction confidence increases from $0.5$ to $0.9$, the ratio decreases almost linearly and the slope stays almost unchanged. But from $0.9$ to $0.99$, the ratio decreases much faster. Based on the observation, we choose the inflection point $0.9$ as the threshold as the trade-off between the number and the quality of selected labels.

In order to further prove our choice, in Table \ref{tab:thre_influ}, we show segmentation results using different thresholds to the self-supervised learning of ${\bf{M}}^{K}_N$ when $K=1$ and $N=1$ in Algorithm \ref{algr:train_proc}. As another option, we also consider soft threshold instead of hard one, namely, every pixel being weighted by its maximum class probability. We show the result on the bottom row. All the results confirm our analysis. When the threshold is lower than $0.9$, the uncorrected prediction becomes the key issue to influence the performance of SSL. While we increase the threshold to $0.95$, the SSL process is more sensitive to the number of pixels that can be used. When we use soft threshold, the result is still worse. It is probably because an amount of labeling noise are involved and the bad impact cannot be well alleviated by assigning a lower weight to the noise label. Thus, $0.9$ seems to be a good choice for the threshold in the following experiments. 

For the iteration number $N$, we select a proper value according to the predicted labels as well. When $N$ increases, the segmentation adaptation model becomes much stronger, causing more labels to be used for SSL. Once the pixel ratio for SSL stops increasing, it means that the learning for the segmentation adaptation model is converged and nearly no improved. We definitely increase the value of $K$ to start another iteration. In Table \ref{tab:N_influ}, we show some segmentation results with the theshold $0.9$ as we increase the value of $N$. We can find the mIoU becomes better with the increasing of $N$. When $N = 2$ or $3$, the mIoU almost stopped increasing, and the pixel ratio stay around the same. It may suggest that $N=2$ is a good choice, and we use it in our work. 

\begin{table*}[!htp]
\scriptsize
\centering
\caption{Comparison results from GTA5 to Cityscapes}
\label{tab:comparison_gta5}
\setlength{\tabcolsep}{3pt}
\begin{tabular}{cccccccccccccccccccccc}
\hline
\multicolumn{22}{ c }{{\scriptsize GTA5 $\rightarrow$ Cityscapes}} \\
\hline
Oracle & {Method}  & \rotatebox{90}{road}  & \rotatebox{90}{sidewalk} &\rotatebox{90}{building} & \rotatebox{90}{wall} & \rotatebox{90}{fence} & \rotatebox{90}{pole} & \rotatebox{90}{t-light} & \rotatebox{90}{t-sign} & \rotatebox{90}{vegetation} & \rotatebox{90}{terrain} & \rotatebox{90}{sky} & \rotatebox{90}{person} & \rotatebox{90}{rider} & \rotatebox{90}{car} & \rotatebox{90}{truck} & \rotatebox{90}{bus} & \rotatebox{90}{train} & \rotatebox{90}{motorbike} & \rotatebox{90}{bicycle} & mIoU\\
\hline
\multirow{5}{*}{\shortstack{ResNet101\cite{he2016deep}\\ 65.1}}
  &Cycada\cite{hoffman2017cycada} & 86.7 & 35.6 & 80.1 & 19.8 & 17.5 & {\bf{38.0}} & {\bf{39.9}} & {\bf{41.5}} & 82.7 & 27.9 & 73.6 & {\bf{64.9}} & 19 & 65.0 & 12.0 & 28.6 & 4.5 & 31.1 & {\bf{42.0}} & 42.7 \\
\cline{2-22}
  &AdaptSegNet\cite{tsai2018learning} & 86.5 & 25.9 & 79.8 & 22.1 & 20.0 & 23.6 & 33.1 & 21.8 & 81.8 & 25.9 & 75.9 & 57.3 & 26.2 & 76.3 & 29.8 & 32.1 & {\bf{7.2}} & 29.5 & 32.5 & 41.4 \\
\cline{2-22}
  &DCAN\cite{wu2018dcan} & 85.0 & 30.8 & 81.3 & 25.8 & 21.2 & 22.2 & 25.4 & 26.6 & 83.4 & 36.7 & 76.2 & 58.9 & 24.9 & 80.7 & 29.5 & 42.9 & 2.50 & 26.9 & 11.6 & 41.7 \\
\cline{2-22}
  &CLAN\cite{luo2018taking} & 87.0 & 27.1 & 79.6 & 27.3 & 23.3 & 28.3 & 35.5 & 24.2 & 83.6 & 27.4 & 74.2 & 58.6 & 28.0 & 76.2 & 33.1 & 36.7 & 6.7 & {\bf{31.9}} & 31.4 & 43.2 \\
\cline{2-22}
  &Ours & {\bf{91.0}} & {\bf{44.7}} & {\bf{84.2}} & {\bf{34.6}} & {\bf{27.6}} & 30.2 & 36.0 & 36.0 & {\bf{85.0}} & {\bf{43.6}} & {\bf{83.0}} & 58.6 & {\bf{31.6}} & {\bf{83.3}} & {\bf{35.3}} & {\bf{49.7}} & 3.3 & 28.8 & 35.6 & {\bf{48.5}} \\
\hline
\multirow{6}{*}{\shortstack{VGG16\cite{simonyan2014very} \\ 60.3}}
  & Curriculum\cite{zhang2017curriculum} & 74.9 & 22.0 & 71.7 & 6.0 & 11.9 & 8.4 & 16.3 & 11.1 & 75.7 & 13.3 & 66.5 & 38.0 & 9.3 & 55.2 & 18.8 & 18.9 & 0.0 & 16.8 & 16.6 & 28.9 \\ 
\cline{2-22}
  & CBST\cite{zou2018unsupervised} & 66.7 & 26.8 & 73.7 & 14.8 & 9.5 & {\bf{28.3}} & 25.9 & 10.1 & 75.5 & 15.7 & 51.6 &47.2 & 6.2 & 71.9 & 3.7 &2.2 & {\bf{5.4}} & 18.9 & {\bf{32.4}} & 30.9 \\
\cline{2-22}
  & Cycada\cite{hoffman2017cycada} & 85.2 & 37.2 & 76.5 & 21.8 & 15.0 & 23.8 & 22.9 & {\bf{21.5}} & 80.5 & 31.3 & 60.7 & 50.5 & 9.0 & 76.9 & 17.1 & 28.2 & 4.5 & 9.8 & 0 & 35.4 \\
\cline{2-22}
  & DCAN\cite{wu2018dcan} & 82.3 & 26.7 & 77.4 & 23.7 & {\bf{20.5}} & 20.4 & {\bf{30.3}} & 15.9 & 80.9 & 25.4 & 69.5 & 52.6 & 11.1 & 79.6 & 24.9 & 21.2 & 1.30 & 17.0 & 6.70 & 36.2 \\
\cline{2-22}
  & CLAN\cite{luo2018taking} & 88.0 & 30.6 & 79.2 & 23.4 & {\bf{20.5}} & 26.1 & 23.0 & 14.8 & 81.6 & 34.5 & 72.0 & 45.8 & 7.9 & 80.5 & {\bf{26.6}} & {\bf{29.9}} & 0.0 & 10.7 & 0.0 & 36.6 \\
\cline{2-22}
  & Ours & {\bf{89.2}} & {\bf{40.9}} & {\bf{81.2}} & {\bf{29.1}} & 19.2 & 14.2 & 29.0 & 19.6 & {\bf{83.7}} & {\bf{35.9}} & {\bf{80.7}} & {\bf{54.7}} & {\bf{23.3}} & {\bf{82.7}} & 25.8 & 28.0 & 2.3 & {\bf{25.7}} & 19.9 & {\bf{41.3}}\\
\hline
\end{tabular}
\vspace{-1em}
\end{table*}
\begin{table*}[!htp]
\scriptsize
\centering
\caption{Comparison results from SYNTHIA to Cityscapes}
\setlength{\tabcolsep}{4.5pt}
\label{tab:comparison_synthia}
\begin{tabular}{ccccccccccccccccccc}
\hline
\multicolumn{18}{ c }{{\scriptsize SYNTHIA $\rightarrow$ Cityscapes}} \\
\hline
Oracle &{Method}  & \rotatebox{90}{road}  & \rotatebox{90}{sidewalk} &\rotatebox{90}{building} &\rotatebox{90}{wall} &\rotatebox{90}{fence} &\rotatebox{90}{pole} & \rotatebox{90}{t-light} & \rotatebox{90}{t-sign} & \rotatebox{90}{vegetation} & \rotatebox{90}{sky} & \rotatebox{90}{person} & \rotatebox{90}{rider} & \rotatebox{90}{car} & \rotatebox{90}{bus} & \rotatebox{90}{motorbike} & \rotatebox{90}{bicycle} & mIoU\\
\hline
\multirow{3}{*}{\shortstack{ResNet101\cite{he2016deep}\\71.7}}
  & AdaptSegNet\cite{tsai2018learning} & 79.2 & 37.2 & 78.8 & - & - & - & 9.9 & 10.5 & 78.2 & 80.5 & 53.5 & 19.6 & 67.0 & 29.5 & 21.6 & 31.3 & 45.9\\
\cline{2-19}
  & CLAN\cite{luo2018taking} & 81.3 & 37.0 & 80.1 & - & - & - & {\bf{16.1}} & {\bf{13.7}} & 78.2 & {\bf{81.5}} & 53.4 & 21.2 & 73.0 & 32.9 & 22.6 & 30.7 & 47.8 \\
\cline{2-19}
  & Ours & {\bf{86.0}} & {\bf{46.7}} & {\bf{80.3}} & - & - & - & 14.1 & 11.6 & {\bf{79.2}}	 & 81.3 & {\bf{54.1}} & {\bf{27.9}} & {\bf{73.7}} & {\bf{42.2}} & {\bf{25.7}} & {\bf{45.3}} & {\bf{51.4}} \\
\hline
\multirow{5}{*}{\shortstack{VGG16\cite{simonyan2014very} \\ 59.5}}
  & FCN wild\cite{hoffman2016fcns} & 11.5 & 19.6 & 30.8 & 4.4 & 0.0 & 20.3 & 0.1 & 11.7 & 42.3 & 68.7 & 51.2 & 3.8 & 54.0 & 3.2 & 0.2 & 0.6 & 20.2 \\
\cline{2-19}
  & Curriculum\cite{zhang2017curriculum} & 65.2 & 26.1 & {\bf{74.9}} & 0.1 & 0.5 & 10.7 & 3.5 & 3.0 & 76.1 & 70.6 & 47.1 & 8.2 & 43.2 & 20.7 & 0.7 & 13.1 & 29.0 \\
\cline{2-19}
  & CBST\cite{zou2018unsupervised} & 69.6 & 28.7 &69.5 & {\bf{12.1}} & 0.1 & {\bf{25.4}} & {\bf{11.9}} & 13.6 & {\bf{82.0}} & {\bf{81.9}} & 49.1 & 14.5 & 66.0 & 6.6 & 3.7 & 32.4 & 35.4 \\
\cline{2-19}
  & DCAN\cite{wu2018dcan} & 79.9 & {\bf{30.4}} & 70.8 & 1.6 &{\bf{0.6}} & 22.3 & 6.7 & 23.0 & 76.9 & 73.9 & 41.9 & 16.7 & 61.7 &11.5 & {\bf{10.3}} & 38.6 & 35.4 \\
\cline{2-19}
  & Ours & {\bf{72.0}} & 30.3 & 74.5 & 0.1 & 0.3 & 24.6 & 10.2 & {\bf{25.2}} & 80.5 & 80.0 & {\bf{54.7}} & {\bf{23.2}} & {\bf{72.7}} & {\bf{24.0}} & 7.5 & {\bf{44.9}} & {\bf{39.0}}\\
\hline
\end{tabular}
\vspace{-1em}
\end{table*}
\section{Experiments}
In this section, we compare the results obtained between our method and the state-of-the-art methods.\vspace{-1.8em}
\paragraph{Network Architecture.}
In our experiments, we choose to use DeepLab V2 \cite{chen2018deeplab} with ResNet101 \cite{he2016deep} and FCN-8s \cite{long2015fully} with VGG16 \cite{simonyan2014very} as our segmentation model. They are initialized with the network pre-trained with ImageNet \cite{krizhevsky2012imagenet}. The discriminator we choose for segmentation adaptation model is similar to \cite{radford2015unsupervised} which has 5 convolution layers with kernel $4 \times 4$ with channel numbers \{64, 128, 256, 512, 1\} and stride of $2$. For each convolutional layer except the last one, a leaky ReLU~\cite{maas2013rectifier} parameterized by $0.2$ is followed. For the image translation model, we follow the architecture of CycleGAN \cite{zhu2017unpaired} with $9$ blocks and add the segmentation adaptation model as the perceptual loss. \vspace{-0.3em}
\paragraph{Training.}
When training CycleGAN \cite{zhu2017unpaired}, the image is randomly cropped to the size $452\times452$ and it is trained for $20$ epochs. For the first $10$ epochs, the learning rate is $0.0002$ and decreases to $0$ linearly after $10$ epochs. We set $\lambda_{GAN}=1$, $\lambda_{recon}=10$ in Equation \ref{equ:trans_loss_D_c} and set $\lambda_{per}=0.1$, $\lambda_{per\_{recon}}=10$ for the perceptual loss. When training the segmentation adaptation model, images are resized with the long side to be $1,024$ and the ratio is kept. Different parameters are used for DeepLab V2 \cite{chen2018deeplab} and FCN-8s \cite{long2015fully}. For DeepLab V2 with ResNet 101, we use SGD as the optimizer. The initial learning rate is $2.5\times10^{-4}$ and decreased with `poly' learning rate policy with power as $0.9$. For FCN-8s with VGG16, we use Adam as the optimizer with momentum as $0.9$ and $0.99$. The initial learning rate is $1\times10^{-5}$ and decreased with `step' learning rate policy with step size as $5000$ and $\gamma=0.1$. For both DeepLab V2 and FCN-8s, we use the same discriminator that is trained with Adam optimizer with initial learning rate as $1\times10^{-4}$ for DeepLab V2 and $1\times10^{-6}$ for FCN-8s. The momentum is set as $0.9$ and $0.99$. We set $\lambda_{adv}=0.001$ for ResNet101 and $1\times10^{-4}$ for FCN-8s in Equation \ref{equ:seg_loss}.\vspace{-0.3em}
\vspace{-1.0em}
\paragraph{Dataset.}
As we have mentioned before, two synthetic datasets -- GTA5 \cite{richter2016playing} and SYNTHIA \cite{ros2016synthia} are used as the source dataset and Cityscapes \cite{cordts2016cityscapes} is used as the target dataset. For GTA5 \cite{richter2016playing}, it contains $24,966$ images with the resolution of $1914\times 1052$ and we use the $19$ common categories between GTA5 and Cityscapes dataset. For SYNTHIA \cite{ros2016synthia}, we use the SYNTHIA-RAND-CITYSCAPES set which contains $9,400$ images with the resolution $1280\times760$ and $16$ common categories with Cityscapes \cite{cordts2016cityscapes}. For Cityscapes \cite{cordts2016cityscapes}, it is splited into training set, validation set and testing set. The training set contains $2,975$ images with the resolution $2048\times 1024$. We use the training set as the target dataset only. Since the ground truth labels for the testing set are missing, we have to use the validation set which contains $500$ images as the testing set in our experiments.   \vspace{-0.3em}
\vspace{-1.0em}
\paragraph{Comparison with State-of-Art.}
We compare the results between our method and the state-of-the-art method with two different backbone networks: ResNet101 and VGG16 respectively. We perform the comparison on two tasks: ``GTA5 to Cityscapes" and ``SYNTHIA to Cityscapes". In Table \ref{tab:comparison_gta5}, we present the adaptation result on the task ``GTA5 to Cityscapes" with ResNet101 and VGG16. We can observe the role of backbone in all domain adaptation methods, namely ResNet101 achieves a much better result than VGG16. In \cite{zhang2017curriculum, tsai2018learning, luo2018taking}, they mainly focus on feature-level alignment with different adversarial loss functions. But working only on the feature level is not enough, even though the best result \cite{wu2018dcan} among them is still about $5\%$ worse than our results. Cycada \cite{hoffman2017cycada} (we run their codes with ResNet101) and DCAN \cite{wu2018dcan} used the translation model followed by the segmentation adaptation model to further reduce the visual domain gap, and both achieved very similar performance. Ours uses similar loss function compared to Cycada \cite{hoffman2017cycada}, but with a new proposed bidirectional learning method, $6\%$ improvement can be achieved. CBST \cite{zou2018unsupervised} proposed a self-training method, and further improved the performance with space prior information. For a fair comparison, we show the results that only use self-training. With VGG16, we can get $10.4\%$ improvement. Therefore, we can find without bidirectional learning, the self-training method is not enough to achieve a good performance.

In Table \ref{tab:comparison_synthia}, we present the adaptation result on the task ``SYNTHIA to Cityscapes" for both ResNet101 and VGG16. The domain gap between SYNTHIA and Cityscapes is much larger than that of GTA5 and Cityscapes, and their categories are not fully overlapped. As the baseline results \cite{tsai2018learning, luo2018taking} chosen for ResNet101 only use $13$ categories, we also list results for the $13$ categories for a fair comparison. We can find from Table \ref{tab:comparison_synthia}, as the domain gap increases, the adaptation result for Cityscapes is much worse compared to the result in Table \ref{tab:comparison_gta5}. For example, the category like `road', `sidewalk' and `car' are more than $10\%$ worse. And this problem will have a bad impact on the SSL because of the lower prediction confidence. But we can still achieve at least $4\%$ better than most of other results given by \cite{zhang2017curriculum,zou2018unsupervised,wu2018dcan,tsai2018learning}.\vspace{-0.3em}
\vspace{-1.5em}
\paragraph{Performance Gap to Upper Bound.}
We use the target dataset with ground truth labels to train a segmentation model, which shares the same backbone that we used, to get the upper-bound result. For ``GTA5 to Cityscapes" with $19$ categories, the upper bounds are 65.1 and 60.3 for ResNet101 and VGG16 respectively. For ``SYNTHIA to Cityscapes" with 13 categories for ResNet101 and 16 categories for VGG16, the upper bounds are 71.7 and 59.5. For our method, although the performance gap is $16.6$ at least, it has been reduced significantly compared to other methods. However, it means there is still big room to improve the performance. We leave it in future work.\vspace{-0.3em}
\section{Conclusion}
In this paper, we propose a bidirectional learning method with self-supervised learning for segmentation adaptation problem. We show via a lot of experiments that segmentation performance for real dataset can be improved when the model is trained bidirectionally and achieve the state-of-the-art result for multiple tasks with different networks.
\section*{Acknowledgment}
This work was partially funded by NSF awards IIS-1546305 and IIS-1637941.

{\small
\bibliographystyle{ieee}
\bibliography{egbib}
}

\end{document}